\documentclass[journal]{IEEEtran}
\usepackage[english]{babel} 
\usepackage{amsmath,amssymb,amsfonts}
\usepackage{graphicx}
\usepackage{caption}
\usepackage{booktabs}
\usepackage{xcolor}
\usepackage{physics}
\usepackage{tikz}
\usepackage{comment}
\usetikzlibrary{quantikz}
\usepackage{blindtext}
\usepackage[ruled,vlined]{algorithm2e}
\usepackage{xcolor}
\usepackage{soul}
\SetKw{KwBy}{by}
\SetKw{KwTo}{to}
\SetKwComment{Comment}{/* }{ */}
\usepackage{float}
\usepackage{balance}
\usepackage{ftnxtra}
\usepackage{fnpos}
\usepackage{caption}
\usepackage{subcaption}
\usepackage{adjustbox}
\usepackage{hyperref}
\usetikzlibrary{shapes.geometric, arrows}

\tikzstyle{startstop} = [rectangle, rounded corners, minimum width=0.3\columnwidth, minimum height=1cm,text centered, draw=white, fill=white]

\tikzstyle{process} = [rectangle, minimum width=0.4\columnwidth, text width = 0.4\columnwidth, minimum height=1cm, text centered, draw=black, fill=orange!60]

\tikzstyle{process2} = [rectangle, minimum width=0.2\columnwidth, text width = 0.2\columnwidth, minimum height=1cm, text centered, draw=black, fill=orange!30]

\tikzstyle{arrow} = [thick,->,>=stealth]

\tikzstyle{textbox} = [rectangle, rounded corners, minimum width=0.1\columnwidth, minimum height=1cm,text centered, draw=white, fill=white]

\tikzstyle{inout} = [rectangle, rounded corners, minimum width=0.1\columnwidth, text width = 0.05\columnwidth, minimum height=1cm,text centered, draw=black, fill=orange!60]

\tikzstyle{gate} = [rectangle, rounded corners, minimum width=0.05\columnwidth, text width = 0.1\columnwidth, minimum height=1cm,text centered, draw=black, fill=white]

\tikzstyle{hidden} = [rectangle, rounded corners, minimum width=0.1\columnwidth, text width = 0.05\columnwidth, minimum height=1cm,text centered, draw=black, fill=red!60]

\tikzstyle{met} = [circle, minimum width=0.05\columnwidth, text width = 0.05\columnwidth, minimum height=1cm,text centered, draw=black, fill=white]

\begin{document}
\bstctlcite{IEEEexample:BSTcontrol}


\title{AI techniques for 
near real-time monitoring\\ of contaminants in coastal
waters\\ on board future $\Phi$sat-2 mission}

\author{Francesca Razzano,~\IEEEmembership{Student Member,~IEEE,} Pietro Di Stasio~\IEEEmembership{Student Member,~IEEE,}\\ Francesco Mauro,~\IEEEmembership{Student~Member,~IEEE,}  Gabriele Meoni,~\IEEEmembership{Member,~IEEE,}~Marco Esposito,~\IEEEmembership{Member,~IEEE,}\\~Gilda~Schirinzi,~\IEEEmembership{Senior~Member,~IEEE}~and~Silvia~L.~Ullo,~\IEEEmembership{Senior~Member,~IEEE}
\thanks{P. Di Stasio, F. Mauro and S. L. Ullo are with the Engineering Department, University of Sannio, Benevento, Italy, email: p.distasio@studenti.unisannio.it, \nobreak f.mauro@studenti.unisannio.it, ullo@unisannio.it} \nobreak 
\thanks{F. Razzano and G. Schirinzi are with Parthenope, Naples, Italy, email: francesca.razzano002@studenti.uniparthenope.it, gilda.schirinzi@uniparthenope.it}
\nobreak \thanks{M. Esposito is with cosine, Netherlands, email: m.esposito@cosine.nl}
\nobreak \thanks{G. Meoni is with University of Delft, Netherlands, email: G.Meoni@tudelft.nl}}

\maketitle
\begin{abstract}
Differently from conventional procedures, the proposed solution advocates for a groundbreaking paradigm in water quality monitoring through the integration of satellite Remote Sensing (RS) data, Artificial Intelligence (AI) techniques, and onboard processing. The objective is to offer nearly real-time detection of contaminants 
in coastal waters 
addressing a significant gap in the existing literature. 
Moreover, the expected outcomes include substantial advancements in environmental monitoring, public health protection, and resource conservation. The specific focus of our study is on the estimation of Turbidity and pH parameters, for 
their implications on human and aquatic health. Nevertheless, the designed framework can be extended to include other parameters of interest in the water environment and beyond.  
Originating from our participation in the European Space Agency (ESA) OrbitalAI Challenge, this article describes the distinctive opportunities and issues for the contaminants' monitoring on 
the $\Phi$sat-2 mission.
The specific characteristics of this mission, with the tools made available, will be presented, with the methodology proposed by the authors for the onboard monitoring of water contaminants in near real-time. Preliminary promising results are discussed and in progress and future work introduced. 
%
\end{abstract}

\section{Introduction}
\IEEEPARstart{T}{he} pressing challenges arising from population growth, escalating water demands for agriculture, energy, and industry, coupled with climate change impacts, underscore the urgent necessity to meticulously monitor and evaluate trends in water resources. This proactive approach ensures the establishment of a solid foundation for water security, guaranteeing sustainable access to safe and usable water. Effective, integrated monitoring of the water cycle's trends and variations, encompassing both quantity and quality, requires combining satellite and in situ observations, data assimilation, and models.
A review of existing observational systems underscores the imperative need for a new, integrated monitoring capability dedicated to water security. The required components for such a capability already exist and can be seamlessly integrated through collaborative efforts among national observational programs \cite{LAWFORD2013633}.
\begin{figure*}[!ht]
\centering
\includegraphics[scale=0.60]{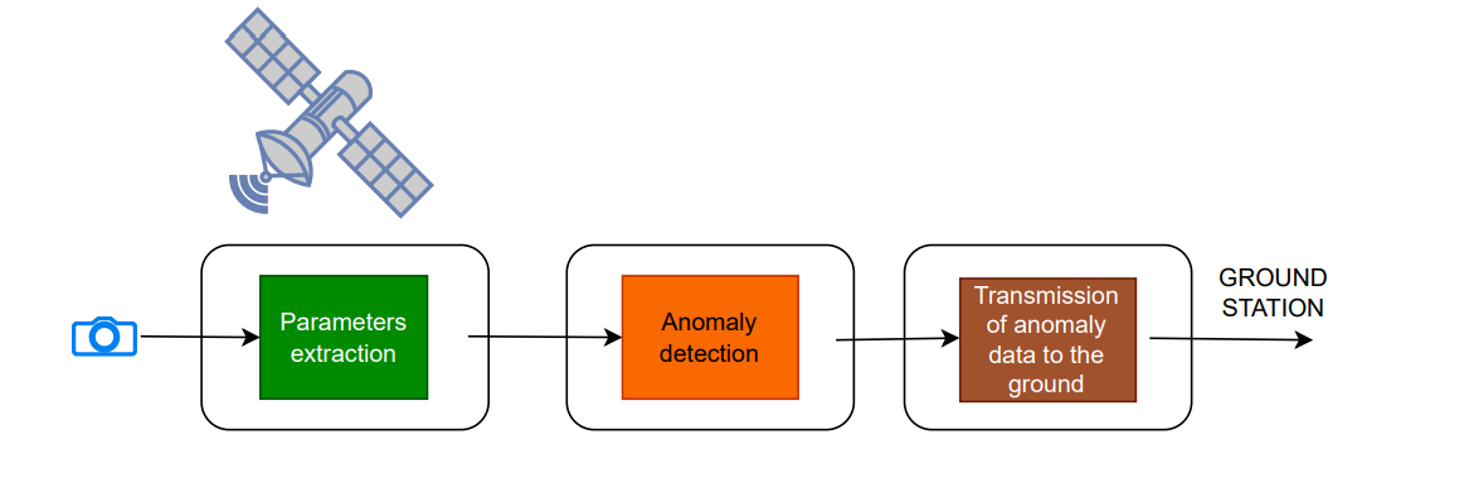}
\caption{Workflow for anomalies' extraction}
\label{system_alert}
\end{figure*}
Furthermore, as the global population grows, particularly in developing countries where problems such as water scarcity and quality concerns are expected to intensify, tensions among different sectors (e.g., agriculture versus urban users) and obstacles to balancing human needs and ecological requirements are inevitable. Tragically, over one and a half million individuals face severe health issues or perish annually due to the lack of access to safe drinking water and sanitation \cite{world2019safer}. Given the escalating pressures on water resources, monitoring assumes critical significance across spatial and temporal scales, providing a systematic and transparent approach to addressing water-related challenges. 
The United Nations and their 2030 Agenda, adopted in September 2015, with the Sustainable Development Goals (SDGs), outline how  among them, 
crucial is to ensure access
\href{https://www.un.org/sustainabledevelopment/water-and-sanitation/}{to water and sanitation for all}. The issue of water pollution presents a considerable challenge to both human health and the environment across numerous countries. Therefore, the importance of monitoring contaminants in water areas cannot be understated, as it is instrumental in seeking effective and meaningful solutions to address both natural and human-induced problems.
The importance of the topic is so remarkable because marine environments suffer from a lack of sufficient sampling, and the prolonged effectiveness of environmental monitoring programs is impeded by inadequate funding. 
Given all the above,
monitoring water contaminants in coastal areas is crucial, as highlighted in numerous scientific articles addressing this issue and emphasizing the main causes and effects for each of them \cite{parameter_nitrate, salinity}. Indicators of water quality, such as its physical, chemical, and biological characteristics, are typically provided by gathering samples in the field and analyzing them in a lab. However, this procedure is impractical, and providing a simultaneous water quality database on a regional scale demands significant labor and time, primarily based on in situ measurements.
Furthermore, conventional point-sample techniques face challenges in capturing the spatial and temporal variations in water quality, crucial for comprehensive assessment and effective waterbody management. This limitation underscores the necessity to explore innovative monitoring methods for detecting contaminants in water. It's worth noting that existing methods leverage Remote Sensing techniques and satellite data. By integrating Remote Sensing techniques with in situ measurements, we can effectively monitor water contaminants. Moreover, satellite data enables the extraction of optical information for a wide range of water quality parameters. One promising approach involves implementing a monitoring process that, relying on predefined acceptable thresholds for water quality parameters, can provide near real-time alerts to emerging risks. This establishes an alert system aimed at safeguarding both human health and the well-being of aquatic ecosystems.
Consequently, there is an increasing demand for smart detection methods that can offer valid solutions to the problem described.\\
The contribution of our work represents an answer to this challenge and aims to realize an innovative method for monitoring near real-time coastal water contaminants, as described ahead in this paper. 
On-board applications may be optimal for early-alert systems, yet there are no studies on onboard marine quality. Compared to terrestrial counterparts, in a properly designated mission, the immediacy of detection can significantly reduce latency in notification \cite{9967616} \cite{thangavel2022near} \cite{del2021board} \cite{furano2020towards}.
The AI model we designed and proposed in this article is engineered to deliver nearly real-time detection of coastal water pollutants through the utilization of satellite data and advanced Machine Learning (ML) algorithms, to be processed on board. Our innovative solution and its monitoring capabilities can be extended to encompass open seas, safeguarding economically vital assets such as fisheries and tourism. This pioneering application of AI and satellite data significantly diminishes response times for pollution detection, offering Governments and decision-makers a tool to ensure the protection of public health.
For the estimation of coastal water pollutants, few works are found in the literature, and in any case, we are the first to propose doing it on board, which represents a major innovation. Our solution adds value to the existing state-of-the-art (SOTA); our onboard AI-powered solution elevates environmental monitoring by furnishing real-time insights, extending coverage, and contributing to a healthier planet.
Existing solutions for water quality monitoring rely on water sampling campaigns and/or involve the use of water quality equipment, leading to substantial operational resources and costs. Simultaneously, conventional approaches are limited to detecting water anomalies after their occurrence, typically when it is too late to implement preventive measures to protect public health or businesses. On the contrary, we aim to find a near real-time solution proposed as shown in Figure \ref{system_alert}, demonstrating how the process involves extracting parameters from an image, identifying anomalies, and transmitting only anomalous data toward the ground stations.

The paper is organized as follows.
The section \ref{section_SOTA} illustrates a wide analysis of the state-of-the-art (SOTA) for the detection of water contaminants, focusing on Artificial Intelligence (AI)-based techniques. In section \ref{section_Paramters}, the main parameters monitored and presented in SOTA are introduced and discussed. The choice of turbidity and pH as the first parameters to study is also justified by their relationship with many other contaminants. A description of the future $\Phi$sat-2 mission is given in section \ref{label_phisat2} and \ref{simulator} to furnish important information on its characteristics to interested researchers, including the details of the Processor and the Simulator that will work on board. In the section \ref{section_implementation_onboard} details on the Myriad 2 device are given. Section \ref{section_datasetcreation} describes the dataset creation, based on in situ data of Ligurian ARPA used as Ground Truth (GT) and the $\Phi$sat-2 satellite data made available through its Simulator.  The proposed methodology is explained in section \ref{label_section_methodology}, highlighting the strengths of onboard AI processing in delivering near real-time solutions for monitoring water contaminants, by filling a gap in the present literature. The models are also introduced in this section with the proposed AI4EDoET (AI for Early Detection of Environmental Threats) architecture and there is also a brief introduction to the metrics. 
Section \ref{sec:results} discusses the results reported for both cases: results with CPU in \ref{sec:resultsA} and results with Myriad 2 \ref{sec:resultsB}. The section \ref{sec:conclusion} ends the paper with discussions and final conclusions.

\section{Background} \label{section_SOTA}
The integration of Remote Sensing (RS) techniques enhances water quality assessment by overcoming traditional limitations, as highlighted in \cite{MOHSENI2022105701}. RS, when combined with in situ measurements, offers significant advantages for monitoring Ocean Water Quality (OWQ) parameters on both large and coastal scales. This approach is based on the optical measurability of most water quality parameters from satellite data. The review covers OWQ parameters such as chlorophyll-a (Chl-a), colored dissolved organic matter (CDOM), Secchi disk depth (SDD), turbidity, total suspended sediments (TSS), sea surface temperature (SST), and chemical oxygen demand (COD), along with monitoring models utilizing various RS datasets. Semi-analytical models coupled with multivariate statistical analysis have gained attention due to their advantages over analytical approaches, which are deemed complex. Empirical and semi-empirical models, though easy to implement, require sufficient in situ measurements for coefficient estimation. Efforts have been made to integrate RS datasets with alternative environmental models to address limitations. Ref. \cite{rs15071938} emphasizes the integration of multispectral sensors for water quality monitoring, particularly beneficial in developing regions. However, challenges arise with coarse-resolution images, necessitating refinement through atmospheric correction models and exploration of high-resolution and hyperspectral imagery. In addition, the evolution of ML models and their integration into RS have introduced a novel dimension to water quality metrics. ML approaches and studies, as \cite{hkong_stations}, \cite{do2023assessing} and \cite{HIBJURRAHAMAN2023138563} are proposed for the analysis and interpretation of WQPs using RS data, aiming to enhance accuracy by detecting complex nonlinear relationships. Research \cite{nazeer2017evaluation} in this context demonstrates that machine learning methods, particularly neural networks (NN), surpass empirical predictive models in estimating SS and Chl-a in complex water bodies. As a result, the study recommends the utilization of machine learning techniques to improve the accuracy and precision of estimating coastal water quality parameters during regular monitoring. With specific analysis, an example is the study conducted by the authors in \cite{VAKILI2020119134} that presents an ANN model to predict optically inactive water quality variables using RS data, achieving superior accuracy compared to linear regression models and by the authors in \cite{47} that employs a RS framework based on a Back-Propagation Neural Network (BPNN) to measure Surface WQPs, though it focuses mainly on local oceanographic processes. \\
Our study introduces AI onboard for contaminant detection, a pioneering approach unmatched in existing literature. This methodology not only accelerates anomaly detection but also offers a forward-looking solution, qualifying for potential deployment in space missions like $\Phi$sat-2 or the International Space Station, following participation in the OrbitalAI Challenge endorsed by ESA $\Phi$-lab.\\
\begin{table*}[Ht]

\caption{Common Turbidity values in natural environments}
\vspace{4pt}
    \centering
    \resizebox{2\columnwidth}{!}{
    \begin{tabular}{c|c}
     
     \small \textcolor{brown}{\textbf{Situation}}  & \small \textcolor{brown}{\textbf{Turbididty value}} \\
     \hline
     \textbf{Standards for drinking water by EPA} &  $<0.3$ NTU in 95\% of samples; never higher than 1 NTU\\
     \hline
     \textbf{Treated water} & 0 to 1 NTU\\
    \hline
     \textbf{Fresh water with visibility exceeding 21.5 inches} & $<10$ NTU\\
    \hline
     \textbf{Fresh water with reduced visibility to 2.5 inches} &  240 NTU\\
    \hline
    \textbf{Brief-term pressure on aquatic organisms}& $>10$ NTU\\
    \hline
    \textbf{Suboptimal conditions for the majority of aquatic organisms} & $>100$ NTU\\
    \hline
    \end{tabular}
\vspace{2pt}
    \label{typical_value}   
    }
\end{table*}

\subsection{Water contaminants parameters}
\label{section_Paramters}
With a focus on key parameters such as Temperature, Dissolved Oxygen, pH, Turbidity, Macroinvertebrates, Escherichia coli (E. coli), Nutrients, Habitat Assessment, and Metals, the rationale behind this monitoring approach is rooted in extensive research, notably drawing inspiration from the U.S. Environmental Protection Agency \href{https://www.epa.gov/}{(EPA)} where the detailed \href{https://www.epa.gov/awma/factsheets-water-quality-parameters}{reasons} are described. In a particular way \textbf{turbidity}, a pivotal parameter, is intricately connected to TSS, functioning as a key indicator of shared water quality characteristics. The impact of elevated turbidity levels on aquatic health is profound, affecting fish gills, visibility for predators, light penetration for aquatic plants, and fish resistance to disease. A variety of factors, both natural and human-induced, contribute to turbidity changes, underscoring the imperative need for continuous monitoring \cite{turbidity_1} \cite{turbidity_2}. Understanding typical turbidity values for \href{https://in-situ.com/us/faq/water-quality-information/what-are-typical-turbidity-values-in-natural-environments}{various scenarios} is paramount as shown in Table \ref{typical_value}. Constant vigilance in monitoring turbidity is crucial for safeguarding water ecosystems and ensuring the well-being of both aquatic life and human populations. For this reason, considering thresholds established by international governmental entities facilitates the alert system process, thereby contributing to mitigating risks to human health.
Ensuring water quality involves vigilant monitoring of various parameters affecting turbidity, with pH standing out as a critical factor. In fact, the logarithmic nature of the pH scale emphasizes its importance, where even a slight one-unit change signifies a ten-fold shift in acidity. pH not only influences water chemistry and toxicity but also affects the solubility and toxicity of metals. Fluctuations in pH levels are a daily occurrence in lakes and rivers, influenced by factors like photosynthesis, respiration of aquatic plants, and human activities and in a specific way we can consider common value of pH as shown in Table\ref{pH_value}.  
\begin{table}[ht!]
\caption{Common pH values in natural environments}
\vspace{4pt}
    \centering
    \resizebox{0.7\columnwidth}{!}{
    \begin{tabular}{c|c}
     \small \textcolor{brown}{\textbf{Situation}} & \small\textcolor{brown}{\textbf{pH value}} \\
     \hline
     \textbf{acidic pH} & $<7$\\
     \hline
     \textbf{Pure water (neutral pH)} &  $7$\\
     \hline
     \textbf{alkaline (basic pH)} & $>7$\\
     \hline
    \end{tabular}
\vspace{2pt}
    \label{pH_value}   
    }
\end{table}

When \href{https://www.fondriest.com/environmental-measurements/parameters/water-quality/ph/}{pH levels} dip below 7.6, coral reefs are prone to collapse due to insufficient calcium carbonate. Freshwater species with heightened sensitivity, like salmon, thrive in pH levels ranging from 7.0 to 8.0. Exposure to levels below 6.0 can result for salmons in severe distress and physiological damage caused by the absorption of metals.\\
The significance of pH monitoring extends to its impact on marine and freshwater ecosystems. In coastal areas, continuous monitoring of turbidity and pH is imperative for safeguarding both human health and the delicate balance of aquatic ecosystems. Furthermore, the inclusion of additional parameters, such as E. coli, adds another layer of importance to water quality assessment. E. coli serves as a vital indicator of fecal contamination, providing insights into the potential presence of disease-causing bacteria and viruses in freshwater. Elevated levels of E. coli pose risks to individuals engaging in recreational activities, leading to symptoms like vomiting and diarrhea \cite{e_coli1} \cite{e_coli2}. The interconnectedness of E. coli concentrations with various parameters like turbidity, TSS, phosphorus, nitrate, and biological oxygen demand (BOD) underscores the comprehensive approach needed for effective WQ management. Monitoring E. coli levels alongside these parameters is crucial for public health and environmental considerations, ensuring the safety of water bodies for recreational activities and maintaining the overall well-being of ecosystems. Having established the crucial role that monitoring contaminants in coastal waters plays in our society, the next section \ref{label_section_methodology} will delve into the methodology and application context of our study, which has yielded promising initial results.

\subsection{$\Phi$sat-2 Mission} \label{label_phisat2}
AI onboard for EO has recently gained a huge interest for the positive impacts on many monitoring applications over our planet. ESA was a pioneer in taking the initial steps in this extremely difficult area of research with the $\Phi$sat-1 satellite launched on September 3, 2020. This was the first of two 6 Units CubeSats that make up the FSSCat mission, also known as the Federated Satellite System (FSSCat).  As regards the second mission,
 \href{https://platform.ai4eo.eu/orbitalai-phisat-2}{the $\Phi$sat-2 satellite}, it will integrate onboard processing capabilities, including AI, along with a Visible to Near Infra-Red (VIS/NIR) multispectral instrument capable of acquiring 8 bands (7 + Panchromatic) at a medium to high resolution. These functionalities will be allocated to a set of specialized applications designed to operate on the spacecraft.

\begin{figure}[!ht]
\centering
\includegraphics[scale=0.40]{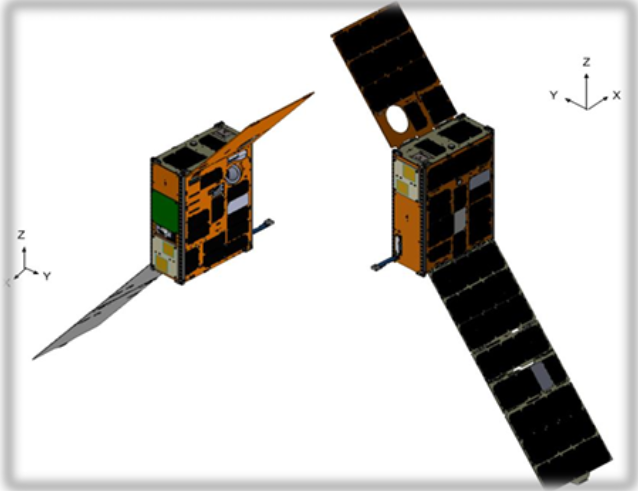}
\caption{
Configuration of $\Phi$sat-2 in the Deployed State}
\label{dips}
\end{figure}

\noindent The satellite is a 6U CubeSat utilizing the established OpenSat 6U platform from Open Cosmos. In Figure \ref{dips} is depicted the $\Phi$sat-2 spacecraft in its deployed configuration.
\begin{table*}[!ht]
 \centering       
  \caption{MultiScape100 Spectral Bands}
  \label{tab:esempio}
\begin{adjustbox}{width=\textwidth}
  \begin{tabular}{c|c|c|c|c|c}  
    
    \small\textcolor{brown}{\textbf{Band}} & \small\textcolor{brown}{\textbf{Centre Wavelength (nm)}} & \small\textcolor{brown}{\textbf{FWHM Bandwidth (nm)}}  & \small\textcolor{brown}{\textbf{HPP Cut-On
(nm)}} & \small\textcolor{brown}{\textbf{HPP Cut-Off(nm)}} & \small\textcolor{brown}{\textbf{Approximate
Sensor Line Number}} \\

\hline
\# 0: PAN & 625 & 250 & 500.0 & 750.0 & 1536\\
\hline
\# 1: MS 1 & 490 & 65 & 457.5 & 522.5 & 2176\\
\hline
\# 2: MS 2 & 560 & 35 & 542.5 & 577.5 & 1964\\

\hline
\# 3: MS 3 & 665 & 30 & 650.0 & 680.0 & 1748\\

\hline
\# 4: MS 4 & 705 & 15 & 697.5 & 712.5 & 1108\\

\hline
\# 5: MS 5 & 740 & 15 & 732.5 & 747.5 & 896\\

\hline
\# 6: MS 6 & 783 & 20 & 773.0 & 793.0 & 680\\

\hline
\# 7: MS 7 & 842 & 115 & 784.5 & 899.5 & 1324\\
    
\hline
  \end{tabular}
  \end{adjustbox}

\end{table*}
\noindent Specifically, concerning the payload aspect, the MultiScape100 instrument developed by Simera Innovate GmbH 
operates as a push-broom imager. This device ensures continuous line-scan imaging across the 8 spectral bands within the VIS/NIR spectral range. The push-broom instrument captures images by scanning along the ground track while the spacecraft orbits the Earth, as illustrated in Figure \ref{fig:payload} below. This scanning process is conducted separately for each spectral band. 

\begin{figure}[!ht]
\centering
\includegraphics[scale=0.50]{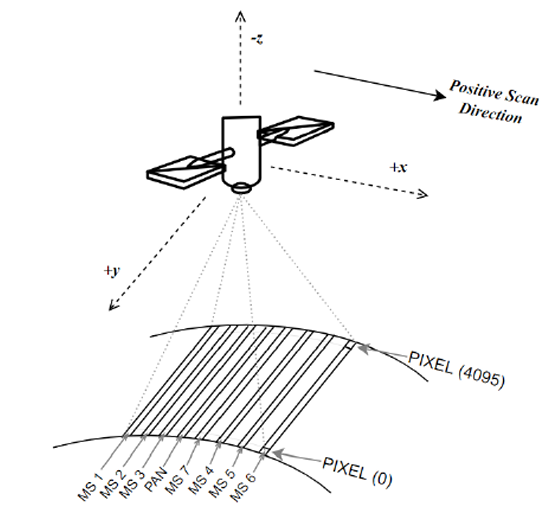}
\caption{Ground Projection of the $\Phi$sat-2 Imager}
\label{fig:payload}
\end{figure}

\noindent Detailed information on each spectral band, including its corresponding line number on the detector plane, is presented in Table \ref{tab:esempio}.

\subsection{$\Phi$sat-2 Simulator} \label{simulator}
Within the framework of the OrbitalAI Challenge, a pivotal simulator has been furnished for $\Phi$sat-2 to assist participants in crafting authentic applications. The primary objective is to provide users with an intuitive tool proficient in realistically simulating diverse products generated on board, unfettered by geographical or temporal coverage constraints. 
The simulator caters to the necessity of simulating any Area of Interest (AOI) without the encumbrance of cost or commercial license restrictions, mandating the utilization of Sentinel-2 data as the primary input. The spectral and spatial attributes of Sentinel-2 furnish a robust underpinning for $\Phi$sat-2, notwithstanding disparities in spatial resolution. 
The spectral bands of $\Phi$sat-2 exhibit a significant higher  spatial resolution (in meters) than Sentinel-2 as shown in Table \ref{tab_value_bands}. In addition, the available products on board $\Phi$sat-2 include those of levels L1A, L1B, and L1C. In our investigation, we employed L1C products, characterized by Top of Atmosphere (ToA) Reflectance in sensor geometry, precise georeferencing, and fine band-to-band alignment (Root Mean Square Error $<$ 10 meters). It is important to note that, in contrast to \href{https://sentinel.esa.int/documents/247904/685211/Sentinel-2-Products-Specification-Document}{Sentinel-2 nomenclature}, \href{https://ai4eo.eu/wp-content/uploads/2023/02/Phisat-2_Mission_Overview_Web.pdf}{$\Phi$sat-2 products} at this level are not orthorectified.

\begin{table*}[!ht]
 \centering       
  \caption{Comparison between the multispectral payloads of Sentinel-2 and $\Phi$sat-2 at a high level, assuming an orbital height of 500km and a consistent Ground Sampling Distance (GSD) of 4.75m across all 8 bands.}
  \label{tab_value_bands}
\begin{adjustbox}{width=\textwidth}
  \begin{tabular}{c|c|c|c|c|c|c|c}
 
  \multicolumn{4}{c|}{\small\textbf{Sentinel-2}} & \multicolumn{4}{c}{\small\textbf{$\Phi$sat-2}}\\
    
    \hline
    \small\textcolor{brown}{\textbf{Id}} & \small\textcolor{brown}{\textbf{Spatial Resolution [m]}} & \small\textcolor{brown}{\textbf{Central Wavelength [nm]}}  & \small\textcolor{brown}{\textbf{Bandwidth [nm]}} &  \small\textcolor{brown}{\textbf{Id}} & \small\textcolor{brown}{\textbf{Spatial Resolution [m]
}} & \small\textcolor{brown}{\textbf{Central Wavelength [nm]}} & \small\textcolor{brown}{\textbf{Bandwidth [nm]}} \\

\hline
2 & 10 & 492.4 & 66 & MS 1 & 4.75 & 490 & 65\\
\hline
3 & 10 & 559.8 & 36 & MS 2 & 4.75 & 560 & 35\\ 
\hline
4 & 10 & 664.6 & 31 & MS 3 & 4.75 & 665 & 30\\
\hline
5 & 20 & 704.1 & 15 & MS 4 & 4.75 & 705 & 15\\
\hline
6 & 20 & 740.5 & 15 & MS 5 & 4.75 & 740 & 15\\
\hline
7 & 20 & 782.8 & 20 & MS 6 & 4.75 & 783 & 20\\
\hline
8 & 10 & 832.8 & 106 & MS 7 & 4.75 & 842 & 115\\
\hline
  &    &       &     & PAN  & 4.75 & 625 & 250\\
\hline
  \end{tabular}
  \end{adjustbox}

\end{table*}

\noindent The operation of the simulator can be summarized as follows: firstly, the Sentinel-2 L1C bands ('B02', 'B03', 'B04', 'B08', 'B05', 'B06', 'B07') and the Scene Classification ('SCL') mask are retrieved, with additional details accessible regarding the bands. Subsequently, separate arrays are generated for the cloud, cloud shadow, and cirrus masks. Optionally, time-frames may be filtered based on data coverage, and metadata concerning solar irradiance and Earth-Sun distance are then acquired. Radiances are computed from reflectances, and a pan-chromatic image is generated through a linear combination of the input bands. Spatial resampling is conducted to align with the pixel size of $\Phi$sat-2 and band-to-band misalignment at the L1A/L1B level is simulated, along with signal degradation attributable to Signal-to-Noise Ratio (SNR) and Module Transfer Function (MTF). For L1C, reflectances are simulated from radiance values, and the AOI is partitioned into a more compact set of image chips. Finally, bands and masks for these image chips are saved, culminating in the creation of an AI-ready dataset.

\section{Myriad 2 device}
\label{section_implementation_onboard}

The AI processing engine integrated into $\Phi$sat-2 is based on the Myriad 2 Vision Processing Unit (VPU). 
The \href{ https://ark.intel.com/content/www/us/ en/ark/products/122461/intel-movidius-myriad-2-vision-processing-unit-4gb.html }{Myriad 2 VPU} is a specialized hardware accelerator designed by Intel/Movidius for handling computer vision and deep neural network tasks. It is part of the Myriad family of VPUs and is specifically tailored for applications such as image and video processing, object recognition, and other vision-related tasks. This multifaceted device, with its compact dimensions, exemplifies a sophisticated and versatile solution for a range of AI  applications \cite{7478823}.\\ 
Several studies investigate the capabilities of Myriad 2, comparing it with other solutions and applying it in various contexts. It has flight heritage with $\Phi$sat-1 and is energy-efficient, with a low development time \cite{rs13081518}. The benefits of using Myriad 2 over traditional field programmable gate arrays and CPUs are assessed by the authors in \cite{leon2021improving}. Other research works, like \cite{9057163}, holds significant importance in this context introducing CNN accelerators using Myriad 2 through two design approaches: deploying CNNs on a power-efficient System on Chip (SoC) and implementing a VHDL application-specific design with a corresponding FPGA architecture. Both systems aim to optimize time performance for specific dataset applications. In current advancements, additional benefits are investigated within the state-of-the-art framework that focuses on enabling an efficient Support Vector Machines (SVM) implementation on an ultra-low-power multi-core SoC, specifically the Intel/Movidius Myriad 2 \cite{8376630}.\\

\section{Dataset creation} 
\label{section_datasetcreation}
\begin{figure}[!ht]
\centering
\includegraphics[scale=0.23]{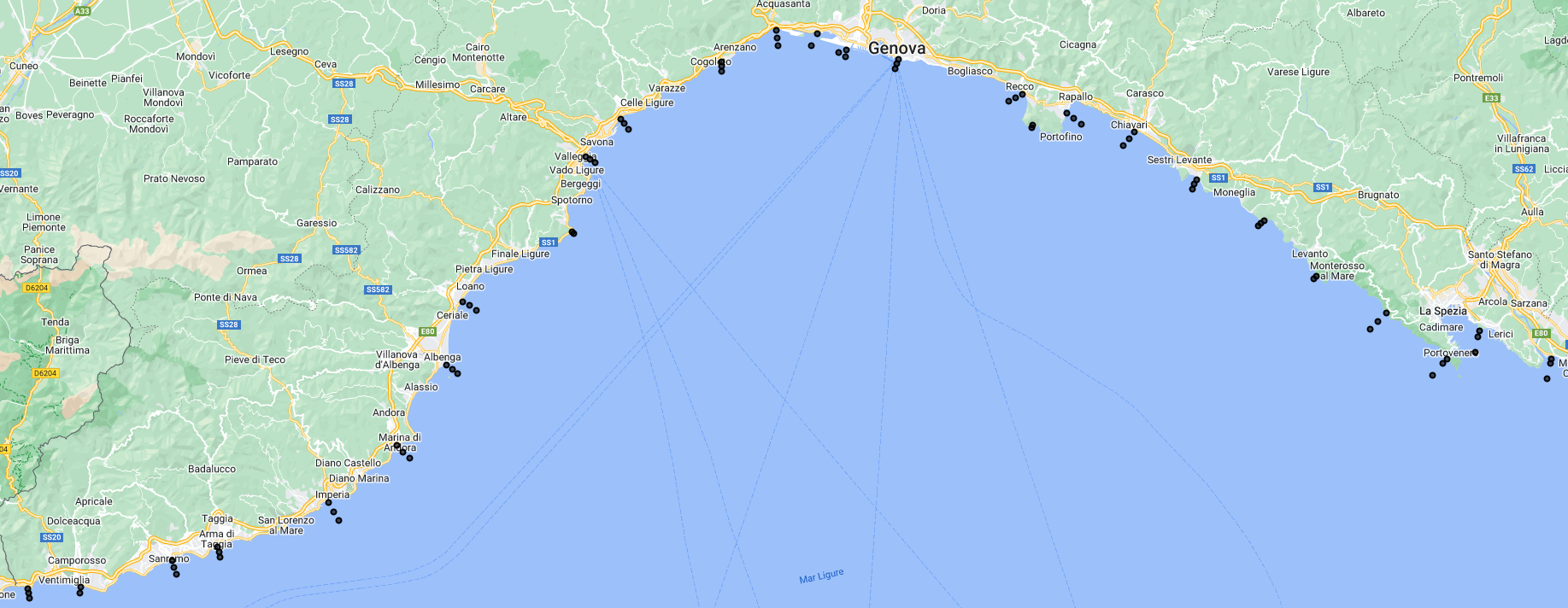}
\caption{Distribution of monitoring station in Liguria region}
\label{fig:map_liguria}
\end{figure}

To create an onboard AI system for the detection of water pollutants using satellite RS data, we built a regression model considering the $\Phi$sat-2 spectral bands as independent variables and the in situ data measurements as dependent variables. In this section we present a suitable dataset created to perform our task. 
For in situ data measurements (chosen as dependent variables), 
an open-access database is made available by 
\href{https://www.arpal.liguria.it/}{ARPA Liguria}
containing measurements of the quality of the Ligurian Sea in waters, sediments and marine organisms for the verification of the health status of the coastal ecosystem and the monitoring of bathing waters. A monitoring network consisting of stable detection points enables retrieval of different parameters, belonging to different environmental matrices, that are periodically analyzed: water, plankton, sediments, benthonic biocenoses. The sampling points, called stations, are identifiable through a description that shows in succession the name of the municipality of belonging, the location name, the distance from the coast and the environmental matrix under investigation. Since 2007, the Liguria coastal waters have been divided into
26 macro-area where the waters, sediments and animal and plant populations are periodically analysed (see Figure \ref{fig:map_liguria}).

For each monitoring station, different dates of acquisition for the parameter under test are present,
and these dates appear repeated several times because, on the same day, different acquisitions were made at different heights of the sensor. In our case, the value of the contaminants has been assessed at the sea surface level to construct the GT. \\
For the creation of the dataset, the first step involved selecting in situ data of to the Liguria region. Subsequently, the identified data were downloaded from the Arpa database. After that, we moved to the simulator Python API and we adjusted the bounding box (bbox) size from 20140 to 1216 meters, creating a square around the Point of Interest (PoI). We considered only a restricted area surrounding the PoI, generating a single patch of dimensions 256$\times$256 pixels (considering a spatial resolution of 4.75 meters). Reducing a patch to 256$\times$256 pixels is also done to accommodate the limited memory requirements of accelerators. It's typical for edge applications \cite{meoni2024ops}. Specifically, $\Phi$sat-2 data temporally and spatially aligned with the in situ data were selected (a temporal tolerance of 3 days has been taken into account, considering the revisit time of $\Phi$sat-2). The implemented code, which will be made available after publication, allows for automated downloading of additional $\Phi$sat-2 data temporally and spatially aligned with new in situ data. The result is a dataset of 1805 unique samples defined by latitude and longitude coordinates, where each of these areas covers a spatial expanse of 1,216 km $\times$ 1,216 km.
\begin{equation}
    X_{lat, lon} \in R^{W \cdot H \cdot B} 
\end{equation}
where \textit{lat} and \textit{lon} are the center coordinates,
\textit{W} the width of the image (256 px), \textit{H} the height of the image (256 px), \textit{B} the number of bands of the image (8).
The data processing level chosen is L1C (without atmospheric correction).

\section{Method proposed}  \label{label_section_methodology}
Since punctual data are available, we started from a Regression model, which is Fully Connected and subsequently, we parallelized it, by transferring the information to a Convolutional Neural Network (CNN) that allowed us to calculate the regression efficiently.
The final network is a CNN that has not been directly trained, but whose weights have been transferred by a regression neural network.
The proposed network has been defined as the AI4EDoET: an AI for Early Detection of Environmental Threats.
The proposed method has been implemented in Python through PyTorch and the code will be made available after publication.

\subsubsection{Fully Connected Neural Network Regression model}
 Our starting AI model is a \textit{Fully Connected Neural Network for regression}, an Artificial Neural Network (ANN) specifically designed to solve regression problems. The goal is to predict pH and turbidity values, given $Phi$sat-2 bands as input. 

 Our model is composed of 5 hidden layers with [512, 512, 512, 512, 43] nodes. Several simulations were performed to set the number of nodes and hidden layers, in such a way as to minimize the root mean square error (RMSE) between the prediction and the GT. \\
Furthermore, the data are systematically prepared,
incorporating a suitable splitting between training, testing,
and validation sets with percentages of 55\%, 20\% and 25\%
respectively. Table \ref{tab:neural_network_details} provides a summary of the layers and configuration details of the neural network.
\begin{table}[!ht]
    \scriptsize
    \centering
    \caption{Fully-connected Neural Network Details}
    \label{tab:neural_network_details}
    \begin{tabular}{c|c}
        
        \small\color{orange}\textbf{Parameter} & \small\color{orange}\textbf{Value} \\
        \hline
        \textbf{Number of hidden layers} & 5 \\
        \hline
        \textbf{Number of nodes in hidden layers} & [512, 512, 512, 512, 43] \\
        \hline
        \textbf{Activation function } & ReLU \\
        \hline
        \textbf{Batch normalization} & Yes \\
        \hline
       \textbf{ Dropout (p)} & 0.25 \\
        \hline
        \textbf{Optimizer} & Adam \\
        \hline
        \textbf{Loss function} & Root Mean Square Error (RMSE) \\
        \hline
        \textbf{Number of epochs} & 2000 \\
        \hline
    \end{tabular}
\end{table}

The models performances are assessed using appropriate regression metrics, such as RMSE and the Mean Absolute Error (MAE).  The choice is also based on the need for consistency and comparability with existing literature or benchmark studies, ensuring a standardized approach to evaluating and communicating the performance of models or methods \cite{hkong_stations} \cite{rs11060617} \cite{rs12071090}. Even though, from the best of our knowledge, there is no presence in the state of the art of an onboard system alert application for water contaminants estimation in coastal waters, we compared our results with ground-based solutions discussed in SOTA.
\\
\subsubsection{AI4EDoET}
Our AI4EDoET network is CNN-based model and processes input image with a shape of 256$\times$256$\times$7. It produces as output a 25$\times$25 matrix. Each output matrix element contains estimates of one water pollutant over 10$\times$10 pixels of the input patch. The first convolutional layer performs spatial averaging over a 10x10 window. In the regressor network, averaging over a 10$\times$10 pixel window was performed as a pre-processing step. In our CNN, this operation is handled by the first layer, which emulates the averaging. It is worth noting that the choice of window size is justified by the fact that initially, on Sentinel-2 data with a spatial resolution of 10 m, averaging over 4 pixels was performed. Now, with PhiSat-2 data, which has a resolution of 4.75 m, we have transitioned to averaging over 10 pixels.
The $n^{th}$ convolutional layer $(n>1)$ maps the $(n-1)^{th}$ fully-connected layer in the corresponding regression network by utilizing 1x1 kernels with no spatial averaging. 
\begin{figure*} [!ht]
\centering
\includegraphics[scale=0.41]{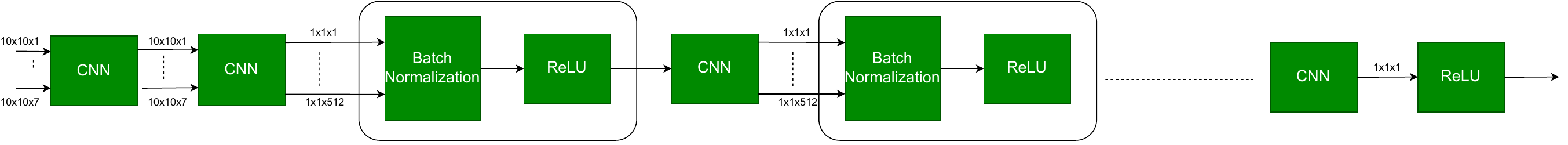}
\caption{AI4EDoET – The proposed Neural Network}
\label{fig:cnn}
\end{figure*}
In Fig. \ref{fig:cnn} it has shown that the network has N+1 layers, where N is the starting regression network layer number, and +1 indicates that the first layer is a convolutional layer that emulates the spatial average made to calculate the data to train this network.
The number of output channels is set equal to the number of neurons in the corresponding fully-connected layer, while the number of input channels matches the number of neurons in the preceding fully-connected layer. Table \ref{net_cnn}
provides a summary of the configuration details of
the neural network.

\begin{table}[!ht]
    \scriptsize
    \centering
    \caption{Convolutional Neural Network Details}
    \label{net_cnn}
    \begin{tabular}{c|c}
        
        \small\color{orange}\textbf{Parameter} & \small\color{orange}\textbf{Value} \\
        \hline
        \textbf{window size} & 10\\
        \hline
        \textbf{input channels} & 7\\
        \hline
        \textbf{hidden layers} & [512, 512, 512, 512, 43]\\
        \hline
        \textbf{output channels} & 1\\
        \hline
    \end{tabular}
\end{table}

We adopted two different models: one for the estimation of pH and one for the Turbidity (the hyparameters setting is the same, but the output weights are different). 

\begin{figure*}[!ht]
\centering
\includegraphics[scale=0.46]{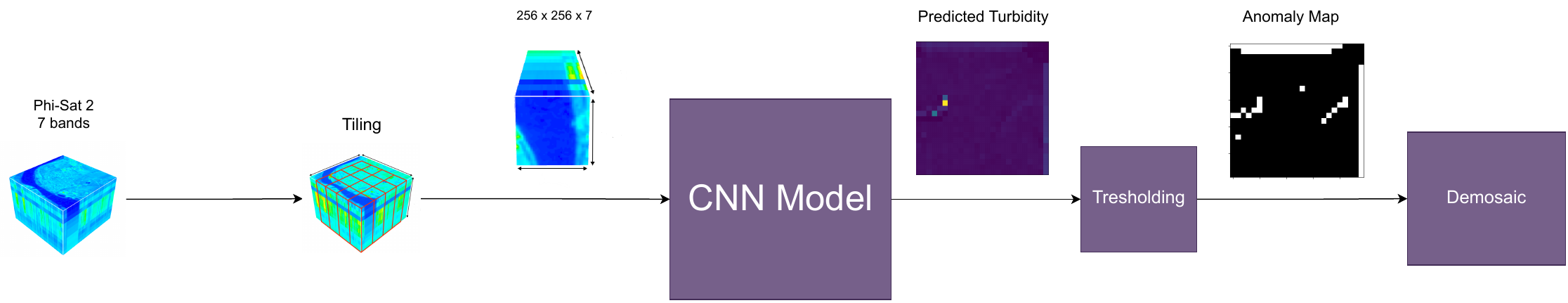}
\caption{Final schematic about the workflow for the implementation of AI4EDoET model onboard for Turbidity estimation}
\label{schematic_onboard}
\end{figure*}
The entire workflow of our proposed approach is schematized in Figure \ref{schematic_onboard}. Since on board, the entire $\Phi$sat-2 tile is available, the initial operation involves tiling the image, resulting in single patches that can be processed by the CNN. Subsequently, binary maps are generated by applying a specified threshold, classifying values as either 0 or 1, based on the alert value of the contaminant. The final step involves reconstruction through a mosaicking process, aimed at restoring the complete input tile of $\Phi$sat-2.
In the context of our partecipation in OrbitalAI Challenge, Ubotica has provided us the access to Rupia, an application designed for remote execution of models on the Myriad 2 device. Rupia introduces a virtual environment tailored for running inference models on Ubotica's Space-ready platforms. This platform allows for remote execution of Edge AI inference without the need for additional hardware. It is supported by a robust and extensively tested backend infrastructure and seamlessly integrates with existing workflows for $\Phi$sat-2.
The implementation on Myriad 2 involved conversion first to ONNX and then through Rupia, with rigorous testing throughout the process

\section{Results} \label{sec:results}
This section is divided into two subsections. The first subsection includes the results obtained for the estimation of Turbidity and pH values using CPU hardware. The second subsection contains the results obtained through Myriad 2 hardware.

\subsection{Results with CPU} \label{sec:resultsA}

\begin{figure}[!ht]
\centering
\includegraphics[scale=0.22]{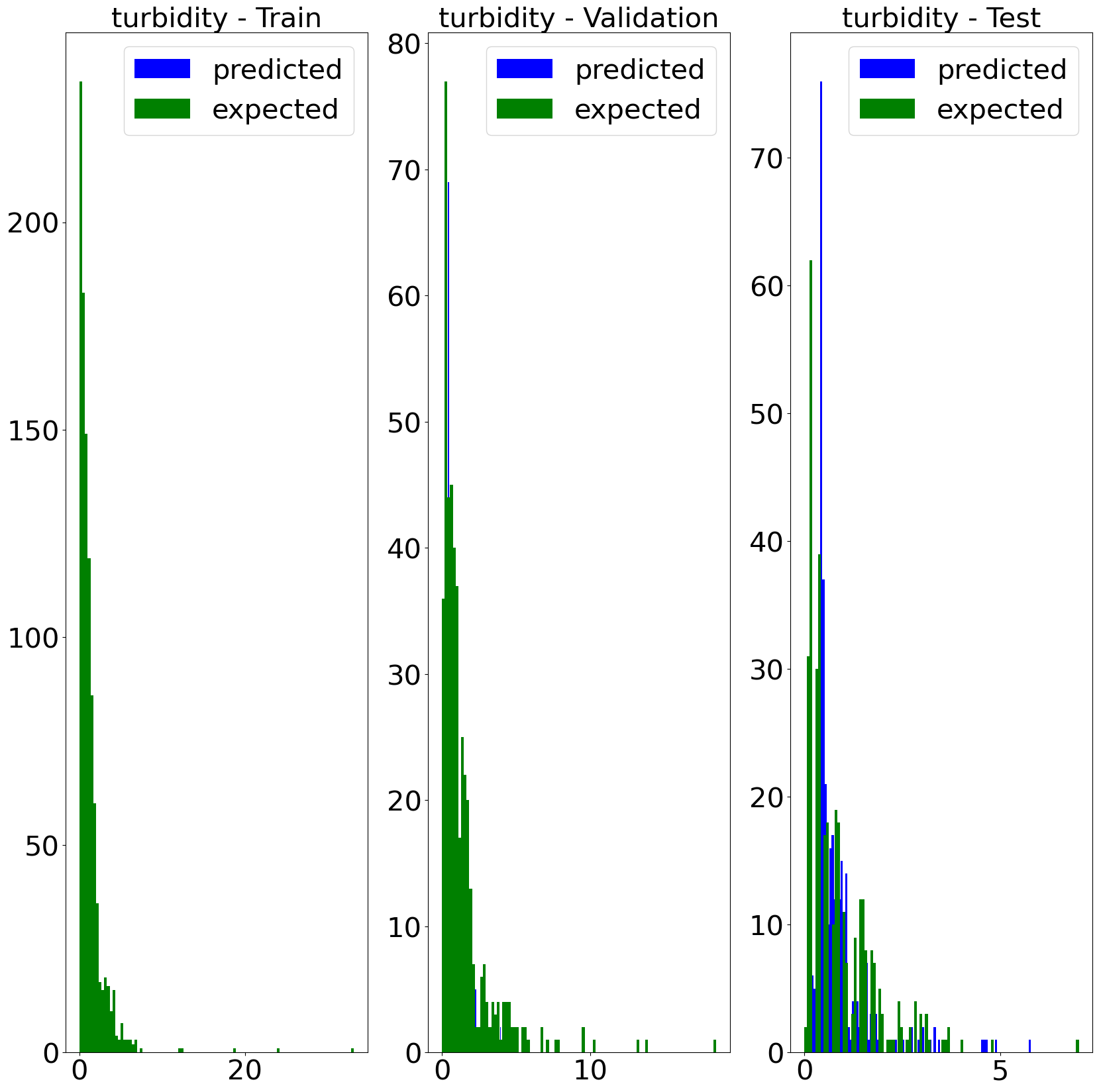}
\caption{Distributions of the prediction and expected Turbidity values}
\label{fig:distribution_Turb}
\end{figure}

\begin{figure}[!ht]
\centering
\includegraphics[scale=0.22]{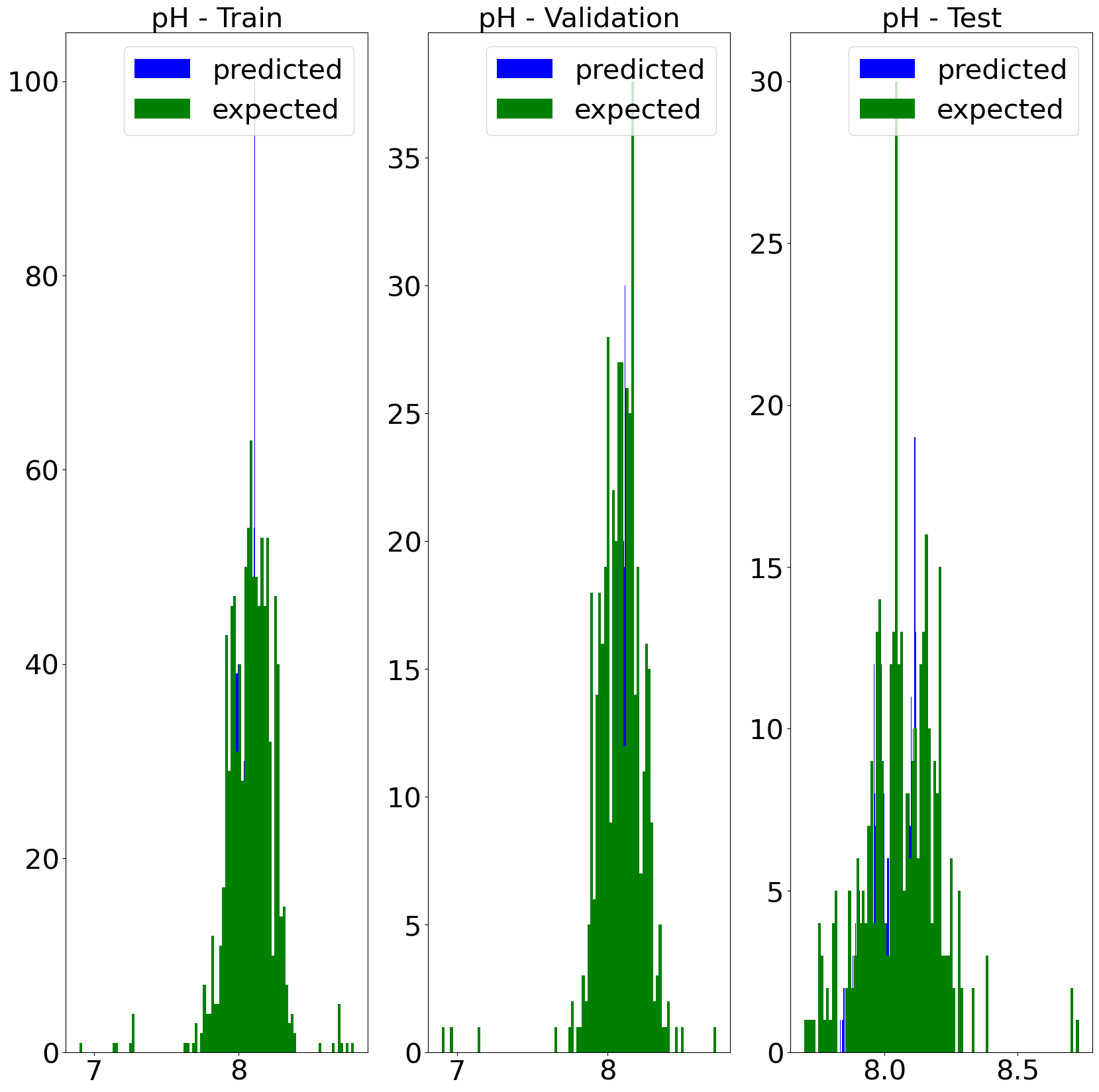}
\caption{Distributions of the prediction and expected pH values}
\label{fig:distribution_pH}
\end{figure}

With regard to the outcomes presented in this section, a qualitative and quantitative analysis is provided concerning Turbidity and pH parameters. We compare the metrics of our results  with other approaches reported in the SOTA that monitor the same contaminants using alternative AI-based frameworks. However, our model not only introduces a valuable innovation in a social context by addressing the issue of onboard water monitoring via satellite in real-time, but it also yields superior results when compared to the mentioned studies against which comparisons are made. It should also be noted that for the proposed model no atmospheric correction of the data is foreseen and this represents a further advantage compared to the other methods present in the SOTA, as the atmospheric correction operation is heavy to do on board \cite{10282605}.
\newline
\noindent In Figure \ref{fig:distribution_Turb} and \ref{fig:distribution_pH} are illustrated qualitative results by depicting the distributions of predictions on the training, validation, and test sets, juxtaposed with the expected distributions of Turbidity and pH parameter (Ground Truth).  As can be seen from the graphs, the estimated distributions map well the trends of the GT distributions.
In a quantitative way, the validation metrics (RMSE and MAE) for Turbidity and pH values are reported in Table \ref{Table_Tur_contaminant} and \ref{Table_pH_contaminant}. These results are compared with other approaches  (\cite{hkong_stations} and \cite{rs11060617}), to show the better performances of our method.

\begin{table}[!ht]
    \scriptsize
    \centering
    \caption{Comparison of results for Turbidity}
    \label{Table_Tur_contaminant}
    \begin{tabular}{c|c|c|c}
        
        \small\color{orange}\textbf{Turbidity [NTU]} & \small\color{orange}\textbf{Methodology} & \small\color{orange}\textbf{RMSE} & \small\color{orange}\textbf{MAE} \\
        \hline
        \textbf{Our study} & \textbf{AI4EDoET} & \textbf{1.0973} & \textbf{0.7426} \\
        \hline
        Study \cite{hkong_stations} & ANN & 1.954 & 1.607\\
        \hline
        Study \cite{rs11060617} & ANN & 3.10 & 2.61\\
        \hline
    \end{tabular}
\end{table}

\begin{table}[!ht]
    \scriptsize
    \centering
    \caption{Comparison of results for pH}
    \label{Table_pH_contaminant}
    \begin{tabular}{c|c|c|c}
        
        \small\color{orange}\textbf{pH} & \small\color{orange}\textbf{Methodology} & \small\color{orange}\textbf{RMSE} & \small\color{orange}\textbf{MAE} \\
        \hline
        \textbf{Our study} & \textbf{AI4EDoET} & \textbf{0.1566} &	\textbf{0.1191}\\
        \hline
        Study \cite{hkong_stations} & ANN & 0.172 & 0.147\\
        \hline
        Study \cite{rs12071090} & SMLR & 0.85 & -\\
        \hline
        Study \cite{rs12071090} & GP model & 0.55 & -\\
        \hline
    \end{tabular}
\end{table}

Making comparisons with pH parameters comparable to the case study proposed in our work has proven challenging from our knowledge. It has been difficult to find an article that addresses a study comparable to ours. This once again highlights the potential of our model and our application, which we propose as an onboard system alert for the upcoming $\Phi$sat-2 mission. It is also important to point out that as regards the feasibility of the model, our model has a size of approximately 4 MB, measured after the conversion to FP 16 used for the Myriad 2, and therefore we are well below the constraint set by ESA of 250 MB.

\subsection{Results with Myriad 2}\label{sec:resultsB}

\begin{figure}
    \centering
    \begin{subfigure}[b]{0.51\textwidth}
        \includegraphics[width=\textwidth]{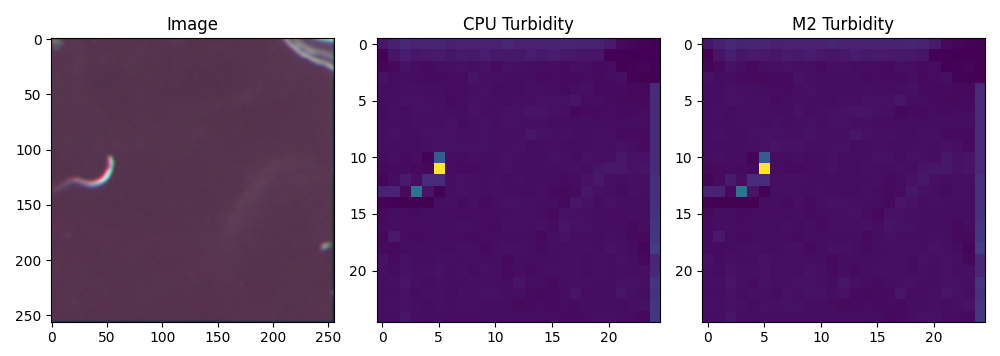}
        \caption{}
    \end{subfigure}
    \hfill
    \begin{subfigure}[b]{0.51\textwidth}
        \includegraphics[width=\textwidth]{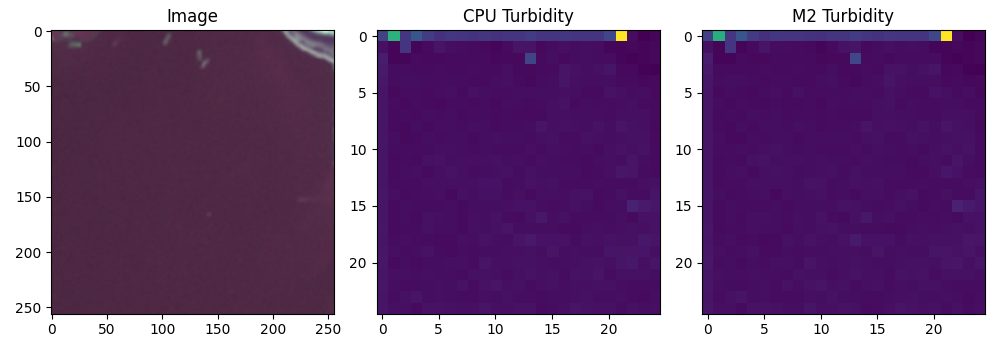}
        \caption{}
    \end{subfigure}
    \hfill
    \begin{subfigure}[b]{0.51\textwidth}
        \includegraphics[width=\textwidth]{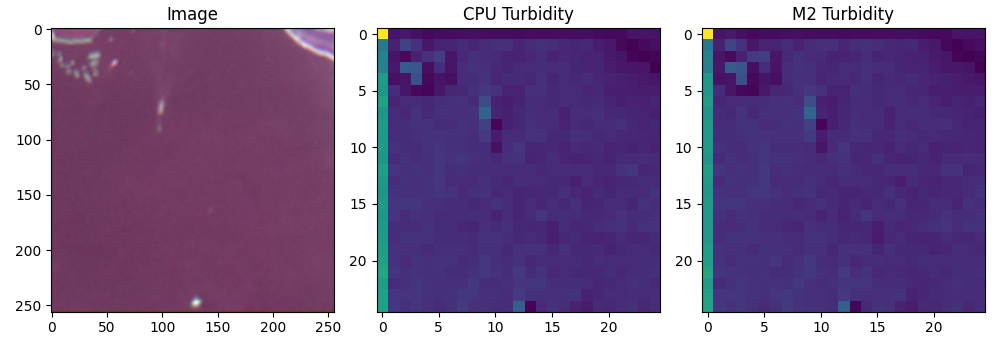}
        \caption{}
    \end{subfigure}
    \caption{Results for final AI onboard-based application}
    \label{fig:example1}
\end{figure}

As this operation is intended to be conducted on board $\Phi$sat-2, to demonstrate the functionality of the inference part on board $\Phi$sat-2, we present in this section some examples of images (in Figure \ref{fig:example1}) computed as the output of our CNN (on our CPU) and the output calculated through the implementation on Myriad 2. The implementation on Myriad 2 involved conversion first to ONNX and then through Rupia, with rigorous testing throughout the process.
From these qualitative results it is clear that we were able to reproduce the same outputs obtained on the CPU also through Myriad 2 and therefore the feasibility of the onboard model was successfully verified. Indeed, in support of the feasibility and compactness of our model, the inference time  is 40.5 ms per inference, with a throughput of 24 FPS.

\section{Discussions and Conclusion}\label{sec:conclusion}
The degradation of both water quality and quantity had been a pressing concern, with various human activities such as residential, agricultural, mining, industrial, power generation, and
forestry operations having negatively impacted the aquatic environment This phenomenon had been particularly pronounced in
coastal areas, where marine contamination had manifested through a multitude of factors affecting the physical, chemical, and bacteriological properties of water, all of which had been linked to intended use and quality standards.
The primary objective of this study has been to employ onboard AI techniques to monitor and detect contaminants in real-time since there are no methods available to do that on board satellite. This proposed approach represents a significant milestone in the context of real-time monitoring of water contaminants. As a result, it can facilitate the rapid generation of alerts and swift interventions when potentially hazardous events are on the horizon. The genesis of this endeavor has been traced back to the OrbitAI Challenge initiated by ESA $\Phi$-lab, with the ultimate goal of deploying an AI-based application on board the
$\Phi$sat-2 mission. There are several solutions that, stemming from this potential study on AI onboard techniques for contaminant monitoring, could be applied for advancements in future research. One such solution is represented by the employment of a different model 
We are currently evaluating various regression models and conducting comparisons to determine if these models yield improved results. Therefore, we are proceeding with a series of analyses involving multiple Machine Learning (ML) and Deep Learning (DL) techniques. The objective is not only to extend the application to additional parameters but also to identify new models capable of performing more accurate simulations. From initial analyses, it is apparent that attention should be given not only to model goodness, assessed through metrics, but also to issues related to implementability on Myriad 2 for onboard compatibility.
Another aspect may be related to expanding the dataset to enhance the network's generalization and yield more accurate results. Furthermore, given the network's capacity to generate valuable predictions for coastal monitoring, it could prove to be a significant asset in regions where acquiring these parameters is challenging. The transferability of the model, in fact, could be one of its major strengths, assisting in contaminant monitoring across various water zones. Another perspective development will entail broadening the monitoring scope to include the collection of chemical contaminants that directly impact human health, such as E. coli. Additionally, future work could focus on comparing our model with others to identify the optimal solution for enhancing this study. In conclusion, this research not only provides valuable insights into coastal water quality monitoring but also underscores the potential for cutting-edge technology and AI to make a meaningful contribution to safeguarding our precious coastal environments and ensuring the well-being of our planet.

\section*{Acknowledgements}
As highlighted in the text, the genesis of this research stems from the participation of the working group in the OrbitalAI Challenge,
endorsed by ESA $\Phi$-lab. All the Team members want to thank the ESA $\Phi$-lab staff and the associated companies who followed us on this unbelievable pathway. 
\bibliographystyle{IEEEtran.bst}
\bibliography{main}
\vspace{-1.0cm}
\begin{IEEEbiography}
[{\includegraphics[width=1in,height=1.15in,clip,keepaspectratio]{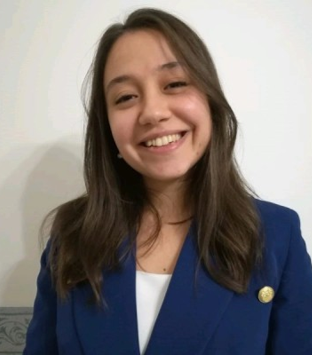}}]{Francesca Razzano} graduated with Laude in Electronic Engineering for Automation and Telecommunications at the University of Sannio in 2023. She is enrolled in the Ph.D. program with University of Parthenope in Information and Communication Technology and Engineering under Supervisor Prof. Gilda Schirinzi, and Co-Supervisor, Prof. Silvia L. Ullo. Her research topics mainly refers to on Remote Sensing and Satellite data analysis, Artificial Intelligence techniques for Earth Observation and data fusion with a particular focus in water quality monitoring. She has co-authored in two papers in reputed conferences for the sector of Remote Sensing. She is a IEEE Student Member with affiliation to the Geoscience and Remote Sensing Society (GRSS) and Aerospace and Electronic Systems Society (AESS). She is a member of a Team competing for OrbitalAI Challenge by ESA-$\Phi$-lab arriving to the last 5 finalist-Team.
\end{IEEEbiography}
\vspace{-1.0cm}
\begin{IEEEbiography}[{\includegraphics[width=1.32in,height=1.18in,clip,keepaspectratio]{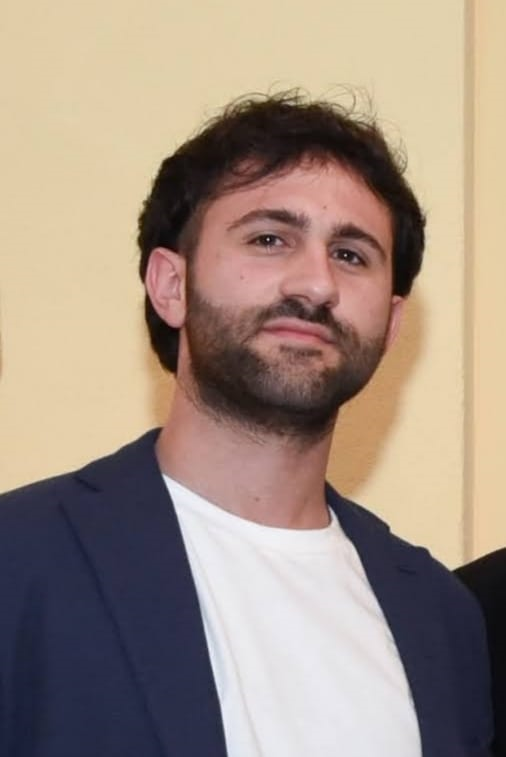}}]{Pietro Di Stasio} Master student at University of Sannio, specializing in Electronics Engineering for Automation and Telecommunications.
His research interests primarily revolve around Earth Observation, Remote Sensing and the application of AI onboard for satellite data analysis.
He was awarded a research scholarship in 2022 at University of Sannio, under Supervisor Prof. Silvia L. Ullo, following his Bachelor's degree.
His work during this scholarship focused on utilizing AI onboard for volcanic eruption detection and he spent about one month as Visiting Researcher at ESA $\Phi$-lab in Frascati, Italy.
He is an IEEE Student Member with affiliation to the Geoscience and Remote Sensing Society (GRSS) and Aerospace and Electronic Systems Society (AESS). He is a member of a Team competing for OrbitalAI Challenge by ESA-$\Phi$-lab arriving to the last 5 finalist-Team. 
\end{IEEEbiography}
\begin{IEEEbiography}[{\includegraphics[width=1in,height=1.15in,clip,keepaspectratio]{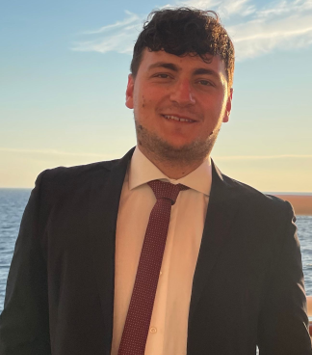}}]{Francesco Mauro} 
Student Member, IEEE received the master degree (with laude and career mention) in Electronic Engineering for Automation and Telecommunications in 2022 from the University of Sannio, Benevento, Italy, where he is currently working toward the Ph.D. degree. He was a Visiting Researcher with the $\Phi$-lab, European Space Agency, Frascati, Italy, and still collaborates with the $\Phi$-lab on topics related to Quantum Machine Learning applied to Earth observation. He has coauthored several papers for the sector of remote sensing and has presented his workd in reputed international conferences. His research interests include remote sensing and satellite data analysis, Artificial Intelligence and Quantum Machine Learning techniques for Earth observation.
\end{IEEEbiography}
\vspace{-1.9cm}
\begin{IEEEbiography}[{\includegraphics[width=1in,height=1.15in,clip,keepaspectratio]{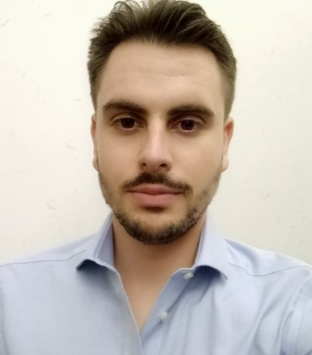}}]{Gabriele Meoni} IEEE Member, received the Laurea degree in electronic engineering from the University of Pisa, Pisa, Italy, in 2016, and the Ph.D. degree in information engineering from the University of Pisa, Pisa PI, Italy, in 2020. During his Ph.D., he developed skills in digital and embedded systems design, digital signal processing, and artificial intelligence. Since 2020, he has been a Research Fellow with the ESA Advanced Concepts Team. His research interests include machine learning, embedded systems, and edge computing. He is currently an Assistant Professor at Delft University of Technology, Delft, Netherlands.
\end{IEEEbiography}
\vspace{-1.8cm}
\begin{IEEEbiography}[{\includegraphics[width=1in,height=1.15in, clip,keepaspectratio]{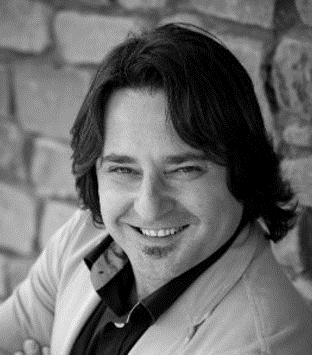}}]{Marco Esposito} received the Ph.D. degree in aerospace engineering specialized in remote sensing completing from the University of Naples, Naples, Italy. Having worked on various aerospace projects in Naples, most recently as a Coordinator of the Environmental Research Aircraft Laboratory of the Italian National Research Council, he moved to cosine Measurement Systems BV, Warmond, Netherlands. Having gained experience as a System and Performance Engineer, a Project Manager, a Program Manager, and a Team Leader at cosine Measurement Systems BV, he became the Managing Director of cosine as well as of cosine Remote Sensing B.V. He has been leading a variety of research and development activities at cosine Measurement Systems, all focusing on the miniaturization of optical instruments for Earth Observation and Planetary Exploration, including NightPod, TropOMI, and Mobile Asteroid Surface Scout(MASCOT) on the Hayabusa 2 Sample Return Mission. He has led the development, demonstration in orbit, and commercialization of the HyperScout®product series, the ﬁrst ever miniaturized imager able to connect hyperspectral, thermal imaging, and artiﬁcial intelligence techniques in one compact product for the Earth Observation Market.
\end{IEEEbiography}
\vspace{-1.1cm}
\begin{IEEEbiography}[{\includegraphics[width=1in,height=1.15in,clip,keepaspectratio]{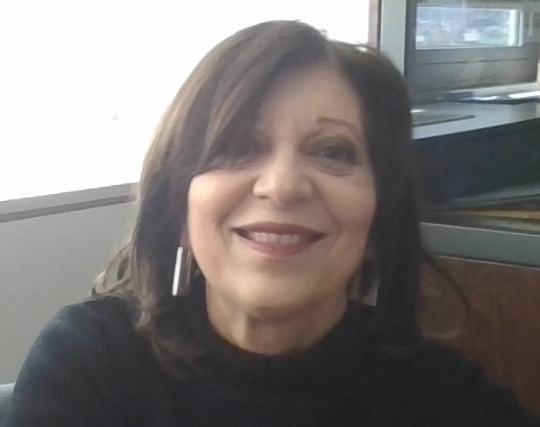}}]{Gilda Schirinzi} graduated cum laude in electronic engineering at the University of Naples "Federico II." From 1985 to 1986, she was at the European Space Agency, ESTEC, Noordwijk, The Netherlands. In 1988, she joined the Istituto di Ricerca per l'Elettromagnetismo e i Componenti Elettronici (IRECECNR), Naples, Italy. In 1998, she joined the University of Cassino, Italy, as an associate professor of telecommunications, and in 2005, she became a full professor. Since 2008, she has been at the University of Naples “Parthenope.” Her main scientific interests are in the field of signal processing for remote sensing applications, with particular reference to synthetic aperture radar (SAR) interferometry and tomography. She is a Senior Member of the IEEE.
\end{IEEEbiography}
\begin{IEEEbiography}[{\includegraphics[width=1in,height=1.15in,clip,keepaspectratio]{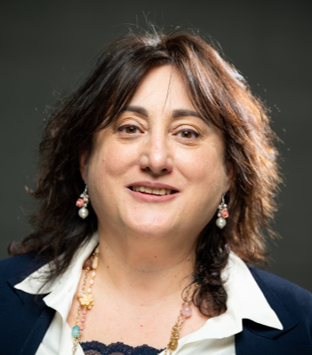}}]{Silvia Liberata Ullo} IEEE Senior Member, President of IEEE AES Italy Chapter, Industry Liaison for IEEE Joint ComSoc/VTS Italy Chapter since 2018, National Referent for FIDAPA BPW Italy Science and Technology Task Force (2019-2021). Member of the Image Analysis and Data Fusion Technical Committee (IADF TC) of the IEEE Geoscience and Remote Sensing Society (GRSS) since 2020. Graduated with laude in 1989 in Electronic Engineering at the University of Naples (Italy), pursued the M.Sc. in Management at MIT (Massachusetts Institute of Technology, USA) in 1992.
Researcher and teacher since 2004 at University of Sannio, Benevento (Italy). Member of Academic Senate and PhD Professors’ Board. Courses: Signal theory and elaboration, Telecommunication networks (Bachelor program); Earth monitoring and mission analysis Lab (Master program), Optical and radar remote sensing (Ph.D. program).  Authored 90+ research papers, co-authored many book chapters and served as editor of two books. Associate Editor of relevant journals (IEEE JSTARS, MDPI Remote Sensing, IET Image Processing, Springer Arabian Journal of Geosciences and Recent Advances in Computer Science and Communications). Guest Editor of many special issues. Research interests: signal processing, radar systems, sensor networks, smart grids, remote sensing, satellite data analysis, machine learning and quantum ML.
\end{IEEEbiography}
\end{document}